\pdfoutput=1

\documentclass[11pt]{article}

\usepackage[]{acl}

\usepackage{times}
\usepackage{latexsym}
\usepackage{booktabs}
\usepackage{multirow}
\usepackage{graphicx}
\usepackage{wrapfig}
\usepackage{subcaption}
\usepackage{gensymb}
\usepackage{bbding}
\usepackage{float}
\usepackage{tikz}

\def\vh{{\bm{h}}}

\def\mD{{\bm{D}}}

\def\mS{{\bm{S}}}
\def\mT{{\bm{T}}}
\def\mU{{\bm{U}}}

\def\mY{{\bm{Y}}}

\def\sR{{\mathbb{R}}}

\usepackage{amsmath,amsfonts,bm}
\usepackage{array}
\newcolumntype{H}{>{\setbox0=\hbox\bgroup}c<{\egroup}@{}}

\newcommand{\test}{\mathcal{D_{\mathrm{test}}}}

\newcommand*\mycirc[1]{%
  \begin{tikzpicture}
    \node[draw,rectangle,inner sep=2.0pt, scale=.8, fill=pink]{#1};
  \end{tikzpicture}}

\newcommand*\myrect[1]{%
  \begin{tikzpicture}
    \node[draw,rectangle,inner sep=2.0pt, scale=.8, fill=green]{#1};
  \end{tikzpicture}}

\def\cmp{\emph{MMP}}
\def\cmpFull{Masked Measurement Prediction}
\def\mnp{\emph{MNP}}

\def\model{\mT}
\def\hOut{\vh_{\model}}
\def\hOutDim{\emph{768}}
\def\hSem{\vh}
\def\hSemDim{\emph{M}}

\newcommand{\isin}[2]{\in \sR^{{#1}\times{#2}}}

\def\context{\mS}
\def\dims{\mD}
\def\setdims{\mathcal{D}}
\def\dim{d}
\def\numDims{\lvert\setdims\rvert}
\def\units{\mU}
\def\setunits{\mathcal{U}}
\def\unit{u}
\def\numUnits{\lvert\setunits\rvert}
\def\num{\mY}

\def\lmae{\emph{log-mae}}
\def\numval{y}

\def\logLaplace{\emph{Log-Laplace}}

\def\mDiscY{\mYDUa \kern .2em \mycirc{-U}\kern .1em \mycirc{-D}} %
\def\mDiscD{\mYDUa \kern .2em \mycirc{-Y}\kern .1em\mycirc{-U}} %

\def\mDiscDU{\mYDUa \kern .2em \mycirc{-Y}} %

\def\mGenYD{\mYDUa \kern .2em \mycirc{-U}} %

\def\mDiscYDU{\mYDUa}

\def\mGenYDU{\mYDUb}

\def\mLatYD{\emph{Lat-Dim}}

\newcommand{\veryshortarrow}[1][3pt]{\mathrel{%
   \hbox{\rule[\dimexpr\fontdimen22\textfont2-.2pt\relax]{#1}{.4pt}}%
   \mkern-4mu\hbox{\usefont{U}{lasy}{m}{n}\symbol{41}}}}
  
\def\mYDUaMeaning{\textbf{Ge}nerative \textbf{M}asked \textbf{M}easurement}
\def\mYDUa{\emph{GeMM}}
\def\mYDUb{\mYDUa \kern .2em \myrect{U$\veryshortarrow$Y}}
\def\numPred{\bar{\mY}}
\def\numTrue{\mY}

\usepackage[T1]{fontenc}

\usepackage[utf8]{inputenc}

\usepackage{microtype}

\title{Masked Measurement Prediction: \\
Learning to Jointly Predict Quantities and Units from Textual Context}

\author{Daniel Spokoyny \\
  Carnegie Mellon University\\
  \texttt{dspokoyn@cs.cmu.edu} \\\And
  Ivan Lee \\
  UC San Diego \\
  \texttt{iylee@ucsd.edu} \\\AND
  Zhao Jin \\
  UC San Diego \\
  \texttt{z3jin@ucsd.edu} \\\And
  Taylor Berg-Kirkpatrick \\
  UC San Diego \\
  \texttt{tberg@ucsd.edu} \\}

\begin{document}
\maketitle

\begin{abstract}
Physical measurements constitute a large portion of numbers in academic papers, engineering reports, and web tables.
Current benchmarks fall short of properly evaluating numeracy of pretrained language models on measurements, hindering research on developing new methods and applying them to numerical tasks.
To that end, we introduce a novel task, \cmpFull\ (\cmp), where a model learns to reconstruct a number together with its associated unit given masked text.
$\cmp$ is useful for both training new numerically informed models as well as evaluating numeracy of existing systems.
In order to address this task, we introduce a new \mYDUaMeaning\ (\mYDUa) model that jointly learns to predict numbers along with their units.
We perform fine-grained analyses comparing our model with various ablations and baselines.
We use linear probing of traditional pretrained transformer models (RoBERTa) to show that they significantly underperform jointly trained number-unit models, highlighting the difficulty of this new task and the benefits of our proposed pre-training approach.
We hope this framework accelerates the progress towards building more robust numerical reasoning systems in the future.
\end{abstract}

\section{Introduction}
Many natural language processing tasks require a deep understanding of numbers -- for example, reading comprehension \cite{Dua2019DROP,ran-etal-2019-numnet}, textual entailment \cite{Sammons2010AskNW,roy2017reasoning} and hybrid table tasks such as fact-verification \cite{2019TabFactA} or question answering \cite{Chen2021FinQAAD}. 
Masked number prediction ($\mnp$) is a popular pretraining objective to imbue language models with numerical understanding and evaluate existing models for their numerical capacity.

As an example of $\mnp$, given the sentence \emph{``Cats have [\#NUM] paws.''} a model learns to predict the number \emph{4}.
While appropriate for numerical commonsense, $\mnp$ is deficient when it is used to predict measurements. 
\emph{Measurements}, such as \emph{2 meters} or \emph{13.2 square miles}, are a special class of particularly common numbers in text that have a well-defined and typed system of \textit{units}.
Given a simple question: \emph{``How long did Alex Honnold climb for?''}, a single number alone is an insufficient answer since it is meaningless without the unit. Answers like \emph{1000 meters} or \emph{4 hours} could both suffice.

\begin{figure}
    \centering
    \includegraphics[width=.50\textwidth]{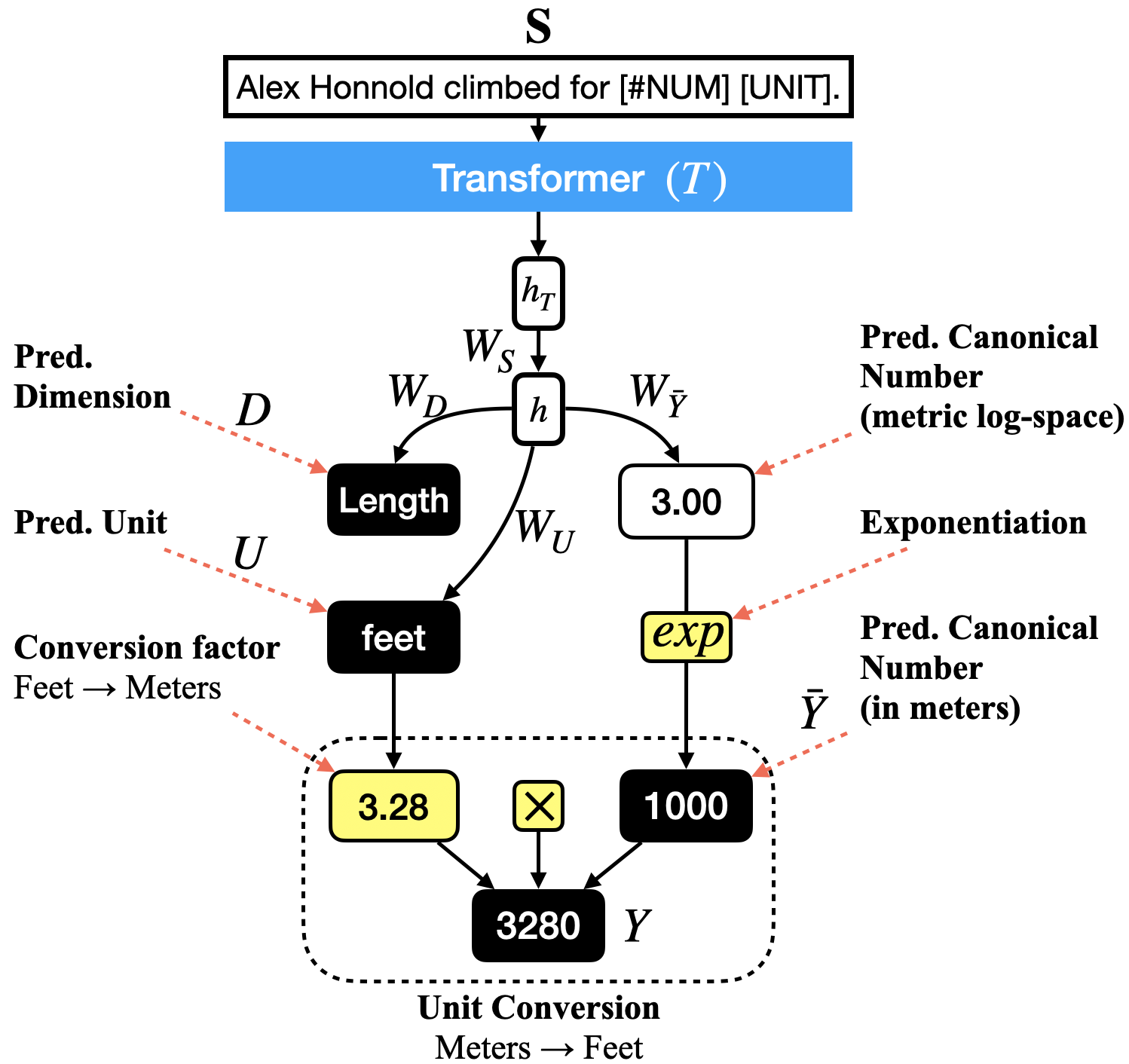}
    \caption{
    We present the \cmpFull\ (\cmp) task where the model predicts the dimension, unit and real-valued number. We also show the model architecture of \mYDUaMeaning\ model (\mDiscYDU), the model we propose to perform \cmp. We display the fixed operations used during unit conversion in yellow.
    In black, we show the different components of the model's prediction.
    }
    \label{fig:task}
\end{figure}

Current $\mnp$ systems do not jointly reason about numbers \emph{with} units.
It is reasonable to expect that pretrained models like BERT could leverage information of units directly as text without any special treatment.
However, in preliminary experiments we find that this yields poor numerical abilities (see Appendix~\ref{app:mlm}).
Furthermore, including units as text directly raise more questions: should we evaluate using all units (\emph{meters, feet, inches})? Should we equally weight across the units? Current models have no opinion about which unit is appropriate because they are not required to make unit predictions during training.
Together, this indicates that current training objectives do not capture sufficient representations of measurements and that a direct application of $\mnp$ to evaluate numeracy of measurements is ill-suited.

To address these shortcomings, we propose the more challenging task of \cmpFull\ ($\cmp$) and a proposed model.
In this task, a model must reconstruct both the number along with the correct units.
In Figure~\ref{fig:task} we show how in a $\cmp$ setting our model generates a dimension (``Length''), a number in metric log-space (``3.00''), the unit ("feet") and then uses the conversion factor (``3.28'') to deterministically output the full measurement (``3280 feet'').
This example illustrates a key distinction in that our model is flexible and can generate \emph{non-metric measurements} (feet) but evaluates numerical prediction in canonical units (meters).\footnote{Our metric of choice described in Equation~\ref{eq-lmae} is invariant to the specific choice of canonical unit i.e., $\lmae$ in meters is equal to $\lmae$ in feet.}

$\cmp$ is useful for two reasons: 1) as a way to \emph{train} models to give them better numeracy
2) as a new kind of \emph{evaluation} that allows for a much more fine-grained analysis of reasoning over numerical quantities.
To perform $\cmp$, we leverage Wiki-Convert \cite{thawani-etal-2021-numeracy}, a large scale dataset of English Wikipedia sentences with ground truth measurement annotations.
We compare the performance of our models on their ability to accurately predict the dimension, unit, and value of a measurement.
We employ a large pretrained transformer model as our textual encoder and examine the performance of different discriminative, generative, and latent variable models along with several ablations.
Our contributions are as follows:
\begin{itemize}
    \item We introduce a novel challenging task $\cmp$ for pretraining and evaluating numeracy.
    \item We show that linear probing of existing pretrained models on MMP \emph{significantly underperforms} fully finetuned models.
    \item We train a model that reasons jointly about numbers and units which predicts numbers 8.1 times more accurately than the probed pretrained models.
    \item On a small-scale human evaluation, we find that our best performing generative model outperforms a set of human annotators achieving 8\% better dimension accuracy and 40\% better unit accuracy. Furthermore, this model predicts a number \textbf{closer} to ground truth 78.75\% of the time compared to our annotators.
\end{itemize}

The task of measurement estimation decouples the different aspects of numeracy allowing for a more interpretable and thorough analysis of numerical reasoning.
Furthermore there are numerous applications of better measurement prediction and unit reconstruction such as in table to text generation \cite{moosavi-SciGen}, answering numerical queries \cite{Sarawagi2014OpendomainQQ, Ho2019QsearchAQ} or for improving comparisons of e-commerce products\cite{Arici2021SolvingPP}.
We hope that \cmpFull\ becomes a standard benchmarking tool from which we can gain insight how to best incorporate new numeracy modeling techniques as well as evaluate existing models.

\section{Models}
\begin{figure}
    \centering
    \includegraphics[width=.4\textwidth]{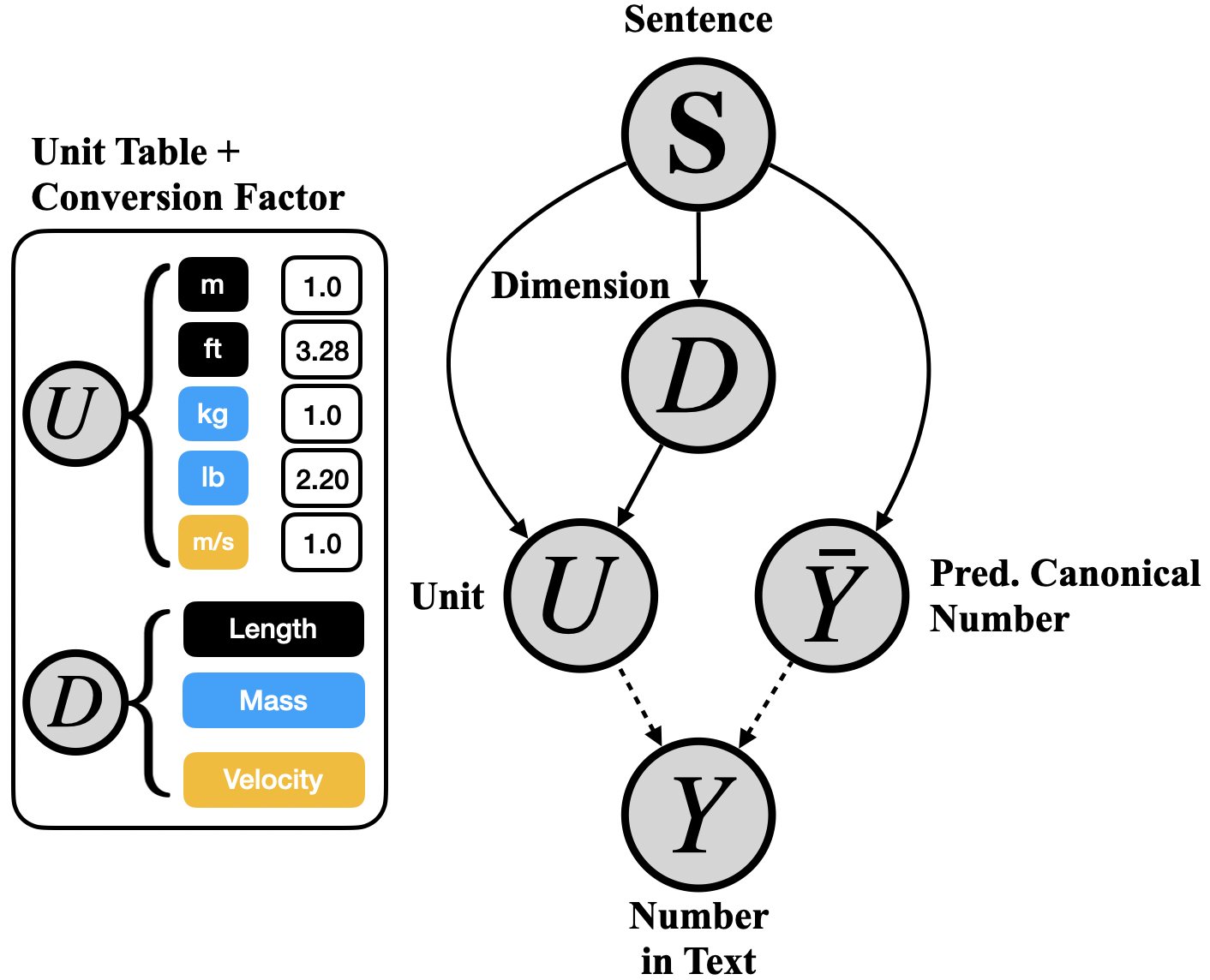}
    \caption{\mDiscYDU\ as a graphical model. The broken arrows represent a deterministic unit conversion. Examples of unit values and their corresponding dimension values are also shown.}
    \label{fig:models}
\end{figure}
\subsection{Background + Notation}\label{background}
The International System of Units (SI) defines seven \emph{fundamental dimensions} (Length, Time, Mass, etc.) and seven corresponding \emph{base SI units} (meters, seconds, kilograms, etc.).
The SI system is the most widely adopted measurement standard and is used internationally in domains such as commerce, finance, logistics, and science.
We designate $\setdims$ to be the set of composite dimensions obtained from (and including) the \emph{fundamental dimensions}. Values of $\setdims$ include velocity and power.
We let $\setunits$ be the set of all units: the various ways to describe dimensions. For example, units of Length include meters and miles. %
Each training example consists of a real number $\numval$, a dimension $\dim \in \setdims$, a unit $\unit \in \setunits$, and the remainder of the sentence $\context$.
In $\cmp$, our task is to predict $\numval$, $\dim$, and $\unit$ given only $\context$. 
In the next sections we describe our generative model designed for $\cmp$ followed by the ablations we consider.

\subsection{Model}
Measurements have complex semantic meanings, shaped by many standards, particular instruments, and natural world phenomena.
Consider a text concerning rainfall. From a dimensional analysis perspective, the units \emph{inches per year (in/y)} and \emph{meters per second (m/s)} share the same dimension \emph{velocity}.
However, mentioning \emph{in/y} usually implies that the text is discussing total rainfall in a region. Likewise, the use of \emph{m/s} suggests that the text is examining the speed of falling rain droplets.
To capture this complexity, we consider a generative model that learns the joint distribution of the number, dimension, and unit.

We now describe the generative process of our full model.
To start, conditioned on $\context$, our model samples a discrete dimension variable $\dims$.
Then conditioned on the sampled dimension, our model samples a discrete unit variable $\units$ compatible with the dimension.
For example, conditioned on the dimension \emph{velocity} our model will output a distribution over the units of velocity such as \emph{[miles per hour; meters per second, inches per year]} as opposed to all of $\setunits$. 
We then separately predict a distribution on the canonicalized measurement, $\numPred$, which is the numerical quantity represented in a base canonical (metric) unit like meters. 
During inference time, we use the highest scoring dimension and unit and choose the proper conversion factor to deterministically produce the final number $\numval$ represented in the predicted unit.
We refer to this \mYDUaMeaning\ model as $\mYDUa$, where the joint $p(\dims,\num,\units|\context)$ is equal to the following equation: 
$$p(\dims|\context) \times p(\units|\dims,\context) \times p(\num|\context)$$
We show the graphical model of $\mYDUa$ in Figure ~\ref{fig:models}.
We also consider, $\mYDUb$, a slight variant where we have a direct dependence between the unit and number prediction such that the joint equals:
$$p(\dims|\context) \times p(\units|\dims,\context) \times p(\num|\units,\context)$$

\subsection{Discrete Latent Dimension Model}
We also consider an unsupervised generative model which treats the dimension as a discrete latent variable.
We use the same number of dimension classes $\numDims$ and train to maximize the log-likelihood of the observed $\num$.
We refer to this model as $\mLatYD$ and is characterized by:
$$p(\num|\context) = \sum_{\dims} p(\dims|\context) \times p(\num|\dims,\context)$$
To evaluate this model we build a contingency matrix of the predicted classes and using a linear solver find the best mapping between our predicted dimensions and true dimensions. We can then apply this mapping to the model predictions and calculate classification metrics for dimension prediction.
\subsection{Model Ablations}
We also consider several model ablations of $\mYDUa$. Our first ablation is $\mDiscD$ which models $p(\dims|\context)$. The second, $\mDiscDU$, learns the distribution $p(\units, \dims|\context) = p(\dims|\context) \times p(\units|\dims,\context)$. The third, $\mGenYD$, models $p(\num, \dims|\context) = p(\dims|\context) \times p(\num|\dims,\context)$.
Our final ablation is $\mDiscY$ which learns $P(\num|\context)$ directly.

\subsection{Model Architectures}
For our textual encoder, we use the Huggingface Transformers \cite{wolf-etal-2020-transformers, liu2019roberta} implementation of RoBERTa, a pretrained 12-layer transformer.
We refer to this text encoder as $\model$ such that given a sentence $\context$, our model outputs a 768-dimensional vector $\hOut$. %
We use a single linear layer, $W_S\isin{\hOutDim}{\hSemDim}$, to project $\hOut$ to $\hSem$ and treat the dimension $\hSemDim$ as a hyper-parameter.
To form a distribution over the real number line $\mathbb{R}$ we use a $\logLaplace$ model, a competitive model used in the numeracy literature \cite{Spokoyny2020AnEI,thawani-etal-2021-numeracy, Zhang2020DoLE}.
This is equivalent to $L_{1}$ regression in log-space and yields the following loss function where $\num$ and $\num^*$ are predicted and ground truth numbers, respectively:
\begin{equation}
    \log{P(\numTrue|\context)} = \left|\log{\num^*} - \log{\num}\right| + \log\left|\frac{1}{\numTrue}\right|
\end{equation}\label{eq:lmae}
As shown in Figure \ref{fig:task}, we project $\hSem$ with a linear layer $W_{\dims}\isin{\hSemDim}{\numDims}$ to obtain a distribution over $\dims$.
We then use a separate linear layer, $W_{\units}\isin{\hSemDim}{\numUnits}$, to project $\hSem$ and obtain a distribution over $\units$.
To predict $\numPred$, we project $\hSem$ with a linear layer $W_{\num}$. 
In the case of \mDiscYDU, we let $W_{\num}\isin{\hSemDim}{\numDims}$ in order to parameterize a mean of a $\logLaplace$ distribution for each dimension in $\dims$.
For \mGenYDU, we set $W_{\num}\isin{\hSemDim}{\numUnits}$ to output the mean of a $\logLaplace$ distribution for each unit in $\units$ and the remaining models, we set $W_{\num}\isin{\hSemDim}{1}$ resulting in a single mean of a $\logLaplace$ distribution.
For training, we use cross-entropy loss for the dimension and unit distributions, and the loss from the equation above for number prediction.

\section{Dataset}
\begin{table}
\centering
\begin{tabular}{lrHHrrH}
\toprule
\textbf{Split} & \textbf{Examples} & \textbf{Neg} & \textbf{Token Size $>$ 64} & \textbf{Max \#} & \textbf{Min \#} & \textbf{\#Char./Tok.}  \\ 
\cmidrule(lr){1-7}
All              & 919,237                            & 5236                   & 34925                               & 5.5E+36     & 1E-06        & 106/33                                  \\ 
\cmidrule(lr){1-7}
Train            & 728,629                            & 3868                   & 27847                               & 5.5E+36     & 1E-06        & XXX                                     \\ 
\cmidrule(lr){1-7}
Val         & 91,110                             & 452                    & 3491                                & 4.4E+14     & 1.2E-06     & XXX                                     \\ 
\cmidrule(lr){1-7}
Test        & 91,092                             & 471                    & 3363                                & 1.6E+21    & 1.8E-06     & XXX                                     \\
\bottomrule
\end{tabular}
\caption{Summary statistics for Wiki-Convert. The median number of characters and tokens per example is $106$ and $33$, respectively.}
\label{tab:data_stats}
\end{table}
\begin{figure*}
\centering
\begin{subfigure}{.5\textwidth}
  \centering
  \includegraphics[width=1\textwidth]{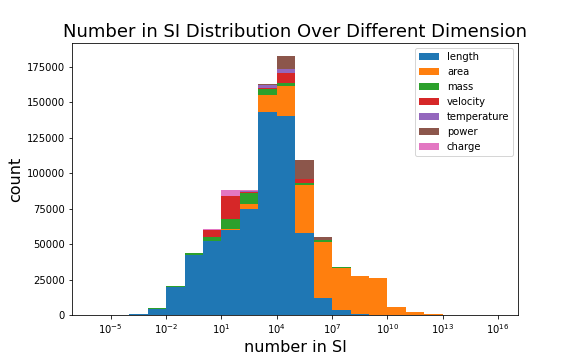}
\end{subfigure}%
\begin{subfigure}{.5\textwidth}
  \centering
  \includegraphics[width=1\textwidth]{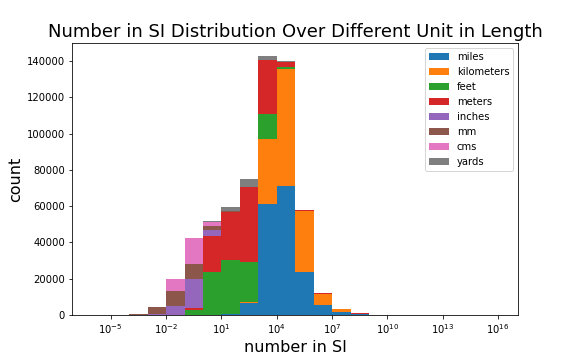}
\end{subfigure}
\caption{Histograms of Wiki-Convert numbers binned by their base-10 exponent. All numbers are canonicalized to their SI form. \textbf{(left)} All numbers labeled by dimension. \textbf{(right)} Numbers that share the \emph{length} dimension labeled by unit.}
\label{fig:data-dist}
\end{figure*}

We train and evaluate our models on Wiki-Convert \cite{thawani-etal-2021-numeracy}, a dataset of English Wikipedia sentences where the number and unit in each sentence are human-annotated.
We canonicalize the units and map each to a single dimension. For example both \emph{feet per second} and \emph{miles per hour} map to \emph{velocity}.
We show the distribution of our all our measurements and all \emph{lengths} in Figure~\ref{fig:data-dist}.
The resulting dataset consists of 919,237 sentences with annotated (number, unit, dimension) triples. 
We provide statistics of the data in Table~\ref{tab:data_stats} with more details in Appendix~\ref{app:data}.
\label{sec:dataset}

\section{Experiments}

We train all models using a batch size of 200 for 100 epochs.
We use the AdamW \cite{Loshchilov2019DecoupledWD} optimizer with a learning rate of $1e^{-4}$ and a linear warm-up schedule of 500 steps.
We use the ``\SnowflakeChevron'' symbol to indicate that we freeze the transformer parameters for training.
For all frozen models we use a log frequency weighted cross-entropy due to the highly imbalanced classes as well as a higher learning rate of $1e^{-3}$.
We employ early stopping with a patience of five epochs on validation score.

To evaluate the performance of our models, we report the macro averaged F1 score for dimension and unit prediction and $\lmae$ to evaluate number prediction. 
We define $\lmae$ in Equation~\ref{eq-lmae} where $\num$ is the predicted number and $\num^*$ is the ground truth number.
As a simple baseline for dimension and unit prediction, we employ majority class voting.
For number prediction we use the median of all the numbers in the training set. 
\begin{equation}\label{eq-lmae}
    \begin{aligned}
    \text{$\lmae$} = \frac{1}{|\test|}\sum_{\test}|\log_{10}\num^* - \log_{10}\num| \\
    \end{aligned}
\end{equation}

\subsection{Few-Shot}
\begin{table}[t]
    \centering
    \resizebox{\columnwidth}{!}{%
    \begin{tabular}{l H H H H H r r r r r }
    \toprule
      \textbf{Model}   & \textbf{Probing Type} & T-10 & T-40 & T-70 & T-100 & \textbf{10-shot} & \textbf{40-shot} & \textbf{70-shot} & \textbf{100-shot} \\
    \midrule
    \mDiscD\SnowflakeChevron & $p(D \vert S)$   & 57.8 & 64.4 & 70.3 & 69.1 & 15.5   & 50.0  & 52.5 & 53.4 \\
    \midrule
    \mDiscD        & $p(D \vert S)$   & 70.4 & 79.3 & 81.6 & 85.2 & \textbf{42.5}  & \textbf{51.2}  & \textbf{57.6} & \textbf{60.5} \\
    \midrule
    Majority       & - & 33.1 & 33.1 & 33.1 & 33.1 & 14.3 & 14.3 & 14.3 & 14.3\\ 
    \bottomrule
    \end{tabular}%
    }
    \caption{Results  of our few-shot experiment on dimension classification (measured by F1 $\uparrow$ and probing $p(D \vert S)$). $x$-shot implies the model is trained on $x$ labeled examples per dimension.
    $\mDiscD$ indicates an ablation of $\mDiscYDU$ where $\num$ and $\units$ are not modeled.
    \SnowflakeChevron\ indicates the model's parameters are frozen during training.
    }
    \label{tab:few_shot_acc_f1}
\end{table}

\begin{table}[t]
    \centering
    \resizebox{\columnwidth}{!}{%
    \begin{tabular}{l H c c c c}
    \toprule
      \textbf{Model}   & \textbf{Probing Type} & \textbf{10-shot} & \textbf{40-shot} & \textbf{70-shot} & \textbf{100-shot} \\
     \midrule
      $\mDiscY$\SnowflakeChevron  &$p(Y \vert S)$& 1.94 & 1.82 & 1.72 & 1.75 \\
      $\mDiscY$                   &$p(Y \vert S)$& \textbf{1.70} & \textbf{1.56} & \textbf{1.43} & \textbf{1.41} \\
     \midrule
      Median &  & 1.99 & 1.99 & 1.99 & 1.99 \\
      \bottomrule
    \end{tabular}
    }
    \caption{Results of our few-shot experiment on number prediction (measured by $\lmae$ $\downarrow$ and probing $p(Y \vert S)$).
    }
    \label{tab:few_shot_lmae}
\end{table}

To study the degree to which current pretrained models capture different aspects of numeracy, we consider the following few-shot experiment.
We sample a balanced dataset of dimensions where each class gets 10, 40, 70, or 100 labeled examples. 
We train $\mDiscDU$ and $\mDiscY$ on the few-shot task where the pretrained text encoder $\model$ parameters are frozen and compare their performance against full fine-tuning.
Due to the high variance of $\mDiscD$, we report the average of three randomly initialized seeds. 
In Table~\ref{tab:few_shot_acc_f1} and Table~\ref{tab:few_shot_lmae} we show results of $\mDiscDU$ and $\mDiscY$ respectively.

As expected, performance improves with more data. However, the frozen models significantly underperform their unfrozen counterparts across all dataset sizes. For example, in the T-100 dataset, the frozen model shows 7.1 lower F1 and 0.34 higher \lmae. These results suggest that current pretrained transformers do not capture numeracy to a large extent.

\subsection{Dimension Prediction}
\begin{table}[t]
    \centering
    \resizebox{\columnwidth}{!}{%
    \begin{tabular}{l l H H r r r}
    \toprule
    \textbf{Model}   & \textbf{Probing Type} & Val & Test &  \textbf{Val} &  \textbf{Test}\\
    \cmidrule(lr){1-6}
    Majority          & -               & 33.1 & 33.1 & 33.1 & 33.1  \\ 
    \cmidrule(lr){1-6}
    $\mDiscYDU$\SnowflakeChevron           & $p(\dims \vert \context)$  & - & -  & 69.1 & 67.5 \\
    \cmidrule(lr){1-6}
    
    $\mDiscD$                               & $p(\dims \vert \context)$  & 97.5 & 97.4 & 88.0 & 86.8 \\
    \cmidrule(lr){1-6}
    $\mDiscDU$                              & $p(\dims \vert \context)$  & 97.4 & 97.5 & 87.0 & \textbf{87.3} \\
    \cmidrule(lr){1-6}
    $\mGenYD$                               & $p(\dims \vert \context)$  & 97.4 & 97.4 & 87.2 & 86.6 \\
    $\mLatYD$                               & $p(\dims \vert \context)$  & - & - & 9.0 & 9.1 \\
    \hline
    \hline
    $\mDiscYDU$                             & $p(\dims \vert \context)$   & 97.4 & 97.5 & 87.4 & 87.0 \\
    \cmidrule(lr){1-6}
    $\mGenYDU$                             & $p(\dims \vert \context)$   & tbd & tbd & 86.4 & 86.1 \\ %
    \bottomrule
    \end{tabular}%
    }
    \caption{Results (\textbf{F1} $\uparrow$) for dimension prediction conditioned on $\context$ only. 
    $\mGenYDU$ indicates a variant of $\mDiscYDU$ where $\numPred$ is dependent on $\units$ (in addition to $\context$).
    }
    \label{tab:dim}
\end{table}

\begin{table}[t]
    \centering
    \resizebox{\columnwidth}{!}{%
    \begin{tabular}{l l H H r r r}
    \toprule
    \textbf{Model}   & \textbf{Probing Type} & Val & Test & \textbf{Val} & \textbf{Test}\\
    \cmidrule(lr){1-6}
    \cmidrule(lr){1-6}
    $\mGenYD$              & $p(\dims \vert \numPred,\context)$ & 98.2 & 98.3 & 95.5 & 95.7 \\
    \cmidrule(lr){1-6}
    
    $\mGenYDU$             & $p(\dims \vert \numPred,\context)$ & tbd & tbd & 96.4 & \textbf{96.6} \\ %
    \bottomrule
    \end{tabular}%
    }
    \caption{Results (\textbf{F1} $\uparrow$) for dimension prediction conditioned on $\numPred$ and $\context$. 
    }
    \label{tab:dim-y}
\end{table}

We train our models and their ablations on the full dataset and measure their performance on dimension prediction.
In Table~\ref{tab:dim}, we show the results of dimension prediction conditioned on $\context$.
We observe that the performance gap between the frozen and unfrozen $\mDiscYDU$ grows to $19.5$ F1 on the test split despite training on 3 orders of magnitude more training data than the few-shot setting.

By using Bayes' rule, we perform dimension prediction conditioned on both $\context$ and $\numPred$ and show our results in Table \ref{tab:dim-y}.
We observe that both models show improved dimension prediction ability when supplied with the number with $\mGenYDU$ reaching 96.6 F1 score, an effective error rate reduction of 75\%.

\subsection{Unit Prediction}
\begin{table}[t]
    \centering
    \resizebox{\columnwidth}{!}{%
    \begin{tabular}{l l H H r r r}
    \toprule
    
    \textbf{Model}   & \textbf{Probing Type} & Val & Test & \textbf{Val} & \textbf{Test}\\
    \cmidrule(lr){1-6}
    Majority                                   & -                & 33.1 & 33.1 & 8.9  & 9.0 \\
     
    \cmidrule(lr){1-6}
    $\mDiscYDU$\SnowflakeChevron                                           & $p(\units \vert \dims,\context)$  & - & - & 29.8 & 29.8 \\
    \cmidrule(lr){1-6} 
    \mDiscDU                                              & $p(\units \vert \dims,\context)$  & 83.2 & 82.9 & 52.9 & 51.7 \\
    \hline 
    \hline 
    $\mDiscYDU$                          & $p(\units \vert \dims,\context)$  & 83.2 & 82.9 & 51.5 & \textbf{54.9} \\                                                          
    \cmidrule(lr){1-6} 
    $\mGenYDU$                           & $p(\units \vert \dims,\context)$  & 83.1 & 82.9 & 49.3 & 47.8 \\                                                
    \bottomrule
    \end{tabular}%
    }
    \caption{Results (\textbf{F1} $\uparrow$) on unit prediction conditioned on the true dimension and text. Ablations are above the double horizontal line.}
    \label{tab:unit}
\end{table}

We show the unit prediction performance of our models in Table~\ref{tab:unit}. The strongest performing model for unit prediction was $\mDiscYDU$ with a F1 score of $54.9$. 
Again, the frozen $\mDiscYDU$\SnowflakeChevron\ produced a $25.1$ lower F1 score than its unfrozen counterpart.

We note that even though the F1 scores on unit prediction are much lower than dimension prediction, they are still significantly better than the majority baseline.
Although one can freely substitute a unit with one in the same dimensional class, we tend to be more systematic and choose units that allow for more straightforward human readability or reflect the actual instruments used for measurement.
As a result, we gravitate towards regularities that models can learn to recognize.
The converse of this is also interesting as it suggests that the expressed units imply more semantic meaning than what is captured in the standardized measurement.

\subsection{Number Prediction}
\begin{table}[t]
    \centering
    \resizebox{\columnwidth}{!}{%
    \begin{tabular}{l l c c}
    \toprule
      \textbf{Model}   & \textbf{Probing Type} & \textbf{Val} & \textbf{Test} \\
     \midrule
      Median                    & -                      & 1.98 & 1.97 \\
      \cmidrule(lr){1-4}
      $\mDiscYDU$\SnowflakeChevron           & $p(\numPred \vert \context)$         & 1.377 & 1.370 \\
      \cmidrule(lr){1-4}
      $\mDiscY$                               & $p(\numPred \vert \context)$         & 0.529 & 0.531 \\
      \cmidrule(lr){1-4}
      \multirow{2}{*}{\mGenYD}              & $p(\numPred \vert \dims,\context)$       & 0.468 & 0.469 \\
                                            & $p(\numPred, \dims \vert \context)$      & 0.517 & 0.518 \\
      \hline 
      \hline
      \rule{0pt}{3ex} $\mDiscYDU$                           & $p(\numPred \vert \context)$         & 0.517 & \textbf{0.515} \\
      \cmidrule(lr){1-4}
      \multirow{2}{*}{\mGenYDU}             & $p(\numPred \vert \units,\dims,\context)$     & 0.401 & \textbf{0.401} \\
                                            & $p(\numPred, \units,\dims \vert \context)$ & 0.526 & 0.526 \\
      \cmidrule(lr){1-4}
      \mLatYD                            & $p(\numPred, \dims \vert \context)$   & 0.545 & 0.546 \\
     \bottomrule
    \end{tabular}
    }
    \caption{ $\lmae$ $\downarrow$ for number prediction conditioned on $\context$. In the second row of $\mGenYD$, we select the highest scoring $\dim^* \in \dims$ and predict $\numval$ conditioned on $\dim^*$ and $\context$. In the second row of $\mGenYDU$, we select the highest scoring $\unit^* \in \units$ and $\dim^* \in \dims$ and predict $\numval$ conditioned on $\unit^*$, $\dim^*$, and $\context$. For $\mLatYD$, we sum over the latent variable $\dims$ to predict $\numval$ conditioned on $\context$.
    }
    \label{tab:numeracy_lmae}
    
\end{table}

We show the number prediction performance of our models in Table~\ref{tab:numeracy_lmae}. Consistent with our previous experiments, all models outperform $\mDiscYDU$\SnowflakeChevron. Furthermore, we observe that not modeling $\units$ and $\dims$ (as is the case in $\mDiscY$) increases $\lmae$, i.e., results in worse numerical prediction. While competitive with $\mDiscYDU$ and its variants on number prediction, $\mLatYD$ cannot predict dimensions with the same efficacy (Table \ref{tab:dim}).

We also experiment with the setting where $\mGenYD$ conditionally generates the number for a particular dimension. In this setting, $\mGenYD$ improves $\lmae$ to $0.469$.
Extending this setting further, we condition $\mGenYDU$ on both a unit and a dimension to produce the best $\lmae$ among our models: $0.401$.

We now revisit our original motivating example: \texttt{Alex Honnold climbed for [NUM] [UNIT]}. Assume we want to know the distance of a climb. To do this, we condition $\mGenYDU$ on $\dims=length$ and $\units=feet$. If, on the other hand, we want to know the duration of a climb, we change the conditioning to $\dims=time$ and $\units=hours$. Now, if we want to know the length of Alex Honnold's climbing career, we condition $\mGenYDU$ on $\dims=time$ and $\units=years$. These examples illustrate the flexibility of $\mGenYDU$ and the importance of jointly modeling numbers, units, and dimensions.

\subsection{Quantitative Analysis}
\subsubsection{Dimensions}

In Figure \ref{fig:d_confmat_gen}, we visualize a confusion matrix of dimension predictions by \mGenYDU.
The low accuracy for electric charge and temperature is attributed to a mislabeling in the dataset.\footnote{Sentences with mislabeled \texttt{Celsius} as \texttt{Coulombs}, which may due to wrong annotation between \texttt{\degree C} and \texttt{C}. Also observed by \citet{Elazar2019HowLA}}
For mass, we find many ambiguous situations where either mass or length are appropriate.
See the first row of Table \ref{tab:human-examples} for such an example.
\begin{figure}
\centering
  \includegraphics[width=.5\textwidth]{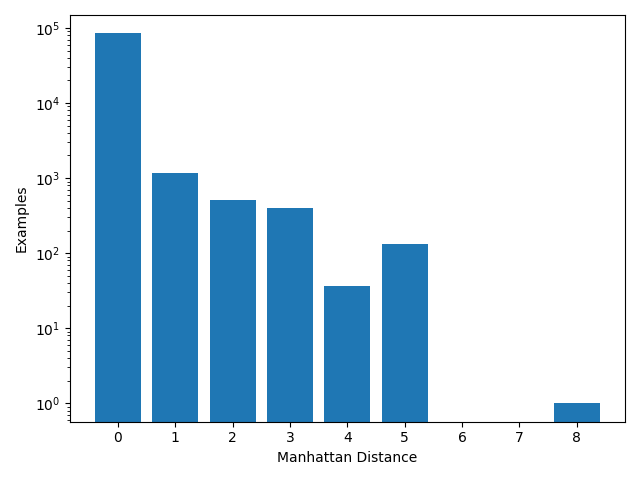}
    \caption{Manhattan distance between true and predicted dimensions by \mGenYDU. 
    We treat dimensions as vectors whose elements are the exponents of the fundamental dimensions that compose a given dimension.
    Note that the y-axis is in log-scale.}
  \label{fig:mandist}
\end{figure}

Thus far, we have treated dimensions as distinct classes with no relationships. However, dimensions are compositions of the seven fundamental dimensions. Therefore, dimensions that share fundamental dimensions are more similar than those that do not. To quantify this similarity, we can treat dimensions as a vector where each element represents the exponent of a fundamental dimension. Then to measure the similarity of two dimensions, we take their Manhattan distance.
To illustrate, assume there exist only two fundamental dimensions: Length and Time. Let $speed = (1,-1)$ and $length = (1,0)$ where the first element represents Length and the second represents Time. The Manhattan distance between $speed$ and $length$ is equal to one.
In Figure \ref{fig:mandist}, we visualize the Manhattan distance between the predictions of $\mGenYDU$ and ground truth.
We observe that there is generally an inverse relationship between error count and the distance of the errors. This observation suggests that our model has learned that some dimensions are more similar than others. This suggestion is reinforced by Figure \ref{fig:d_confmat_gen} where misclassifications tend to have small distances from the true dimension. For example, velocity is most often misclassified as length.

\subsubsection{Units}

\begin{figure}[h!]
\centering
\begin{subfigure}{.5\textwidth}
    \centering
    \includegraphics[width=.98\textwidth]{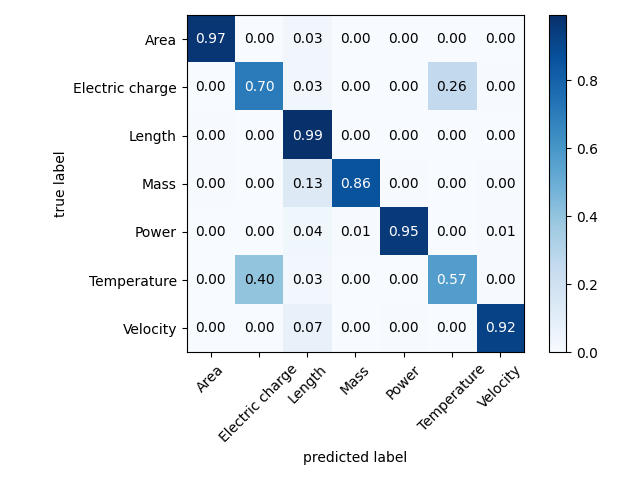}
    \vspace{-1em}
    \caption{}
    \label{fig:d_confmat_gen}
\end{subfigure}\hfill
\begin{subfigure}{.5\textwidth}
    \centering
    \includegraphics[width=.98\textwidth]{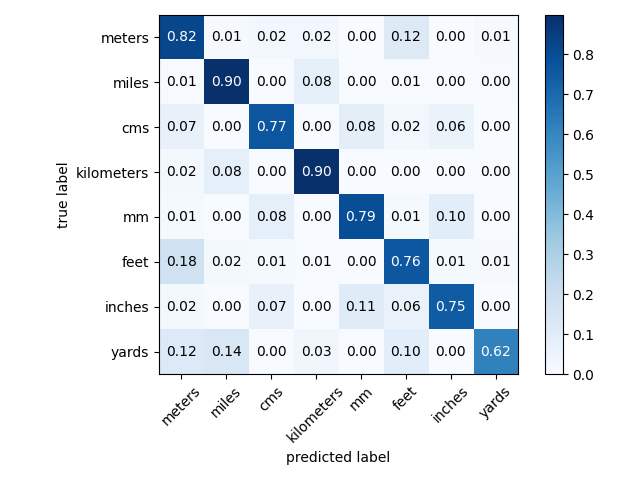}
    \vspace{-1em}
    \caption{}
    \label{fig:u_confmat_gen}
\end{subfigure}\hfill
\caption{Confusion matrices for predictions by $\mGenYDU$ over the validation split.
\textbf{(top \ref{fig:d_confmat_gen})} Dimension prediction. Most misclassified dimensions are similar to their ground truth counterparts in terms of Manhattan distance.
\textbf{(bottom \ref{fig:u_confmat_gen})} Unit prediction for examples that share the \emph{length} dimension. Most misclassified units of length share similar magnitudes to their ground truth units.}

\label{fig:confusion_m}
\end{figure}

In Figure \ref{fig:u_confmat_gen} we the show the confusion matrix on unit predictions for \emph{lengths}.
We find that most mistakes occur substituting units with ones that have similar magnitudes like feet for meters or kilometers for miles.
The model struggled predicting yards possibly due to the lower number of examples (Figure~\ref{fig:data-dist}).

\subsubsection{Numeracy}
\begin{table}[t]
    \centering
    \resizebox{\columnwidth}{!}{%
    \begin{tabular}{cccccc}
    \toprule
    \textbf{Length}  & \textbf{Area} & \textbf{Velocity} & \textbf{Mass} & \textbf{Power}\\
    \midrule
    0.37 & 0.54 & 0.19 & 0.55 & 0.27\\
    \bottomrule
    \end{tabular}
    }
    \caption{$\lmae$ $\downarrow$ by dimension. Numbers for some dimensions such as Area and Mass are more difficult to predict than others.}
    \label{tab:lmae-dim}
\end{table}
In Table~\ref{tab:lmae-dim}, we show $\lmae$ by dimension as predicted by $\mGenYDU$.
We note that errors are not uniform across dimensions, predicting \emph{areas} is 2.2 times harder \emph{velocities}.
We also observe that the magnitudes of errors seem somewhat to be positively correlated with the variances observed in Figure~\ref{fig:data-dist}.

\subsubsection{Human Evaluation}\label{sec:human}
\begin{table}
\centering
\resizebox{\columnwidth}{!}{
\begin{tabular}{lrrrrrrrrr}
\toprule
      & \multicolumn{2}{c}{\textbf{Model}} & \multicolumn{2}{c}{\textbf{Human}} & \textbf{Model >}\\
      & & & & & \textbf{Human \ }\\
  \cmidrule(lr){2-3} \cmidrule(lr){4-5} \cmidrule(lr){6-6}
   & $\dims$ & $\units$ & $\dims$ & $\units$ & $\num$ \\
   \cmidrule(lr){1-6}
   \textbf{Cor.} & 87 &75 &80 &37 & 63\\
   \cmidrule(lr){1-6}
   \textbf{Cnt.} & 90 & 87& 90 & 80 &80\\
   \cmidrule(lr){1-6}
   \textbf{Acc.} & \textbf{96.7}&	\textbf{86.2}&88.9& 46.3& \textbf{78.8}\\
  
\bottomrule
\end{tabular}
}
\caption{Dimension and unit prediction accuracy of our human evaluation experiment. $\mGenYDU$ prediction on dimension and unit both surpassed human performance.
The final column shows that the model predicted a number closer to ground truth in $78.8\%$ of the cases.}
\label{tab:human-tab}
\end{table}

\begin{table*}[t]
    \centering
    \resizebox{\textwidth}{!}{%
    \begin{tabular}{p{0.01\linewidth}p{.5\linewidth}p{0.08\linewidth}p{0.08\linewidth}p{0.08\linewidth}p{0.08\linewidth}p{0.08\linewidth}p{0.08\linewidth}p{0.08\linewidth}p{0.08\linewidth}p{0.08\linewidth}}
    \toprule
    
    & & \multicolumn{3}{c}{\textbf{True}} & \multicolumn{3}{c}{$\mGenYDU$ \textbf{Prediction}} & \multicolumn{3}{c}{\textbf{Human Prediction}}\\
    
    \cmidrule(lr){3-5} \cmidrule(lr){6-8}\cmidrule(lr){9-11}
    \# & \textbf{Text} & Dim & Unit & Num & Dim & Unit & Num & Dim & Unit & Num \\
    \midrule
    1 & Hope is gaff rigged, 'V'-bottomed and has an [\#NUM] [\#UNIT] centerboard. & Mass & pounds & 385.554 & Length & feet & 2.971  & Length & meter & 50
    \\
    \midrule
    2 & Some have been running for over 50 years, each covering about [\#NUM] [\#UNIT]. & Velocity & miles per year & 0.102 & Area & sqkm & 2.09E+10  & Area & sqmi & 2.59E+07 \\
    \midrule
    3 & Another medium-sized corvid, the [\#NUM] [\#UNIT] Eurasian magpie (Pica pica) is also amongst the most widely reported secondary prey species for goshawks there. & Mass & grams & 0.218 & \textbf{Mass} & \textbf{grams} & 0.049  & \textbf{Mass} & \textbf{grams} & \textbf{0.2}\\
    \midrule
    4 & The twin cylinder, liquid-cooled, in-line two-stroke, [\#NUM] [\#UNIT] Rotax 582 has also been used. & Power & horse-power & 47725 & \textbf{Power} & \textbf{horse-power} & 39248  & \textbf{Power} & \textbf{horse-power} & \textbf{45000}\\
    \midrule
    5 & Chrysothamnus may grow up to a [\#NUM] [\#UNIT] tall shrub or subshrub, usually with woody stem bases & Length & cms & 1.2 & \textbf{Length} & meters & \textbf{1.147}  & \textbf{Length} & meters & 1\\
    \midrule
    6 & Kurt Busch was the fastest in the first practice session with a time of 21.372 seconds and a speed of [\#NUM] [\#UNIT]. &  Velocity & mph & 75.103 &\textbf{Velocity} & \textbf{mph} & \textbf{63.584} & \textbf{Velocity} & meters per second   & 10 \\
    \bottomrule
    \end{tabular}%
    }
    \caption{Instances of the $\cmp$ task performed during our human evaluation experiment, all numbers are in SI units. 
    In example 1, both the model and humans all predict the incorrect  dimension length instead of mass. 
    The preceding sentence of example 2 references travelling trains leading both to incorrectly predict area instead of velocity. 
    In example 6 the model predicts the speed of the NASCAR driver Kurt Busch's car
    whereas the humans had mistaken him for a runner.
    }
    \label{tab:human-examples}
\end{table*}
We compare the $\mGenYDU$ against the combined effort of three average annotators on a balanced set of 90 sentences sampled randomly from the test set. The three annotators have diverse scientific backgrounds ranging from chemistry, earth sciences, and computer science. One annotator is a native Chinese speaker, and two are native English speakers. The annotators worked together to predict the missing dimensions, units, and accurate measurement estimates. We show examples of the sentences and their predictions in Table~\ref{tab:human-examples} and the results in Table~\ref{tab:human-tab}. We find that the model outperforms the human annotators on every task. For dimension prediction, the model led by 8 percentage points. Of the sentences where the dimension was correctly classified, the model led by 40 percentage points on unit prediction. For sentences where both the model and human correctly predicted the dimension, the model predicted a number closer to ground truth 79\% of the time.

\subsection{Qualitative Analysis}

\subsubsection{Semantic Head Embeddings}
\begin{figure*}[h!]
\centering
\begin{subfigure}{.33\textwidth}
    \centering
    \includegraphics[width=1\textwidth]{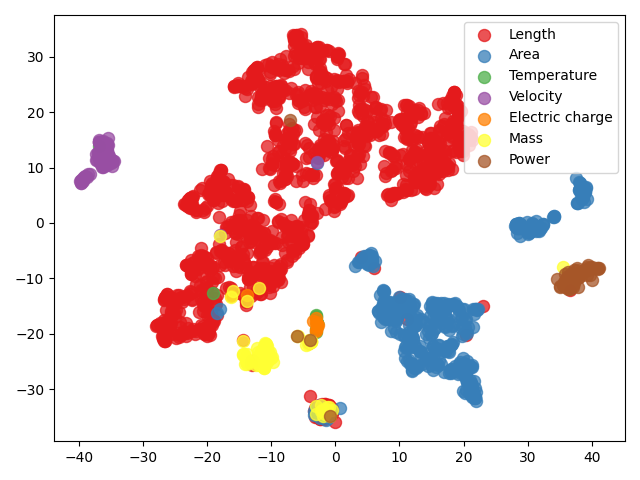}
    \hspace*{\fill}%
    \vspace{-2em}
    \caption{}
    \label{fig:d_tsne}  
\end{subfigure}%
\begin{subfigure}{.33\textwidth}
    \centering
    \includegraphics[width=1\textwidth]{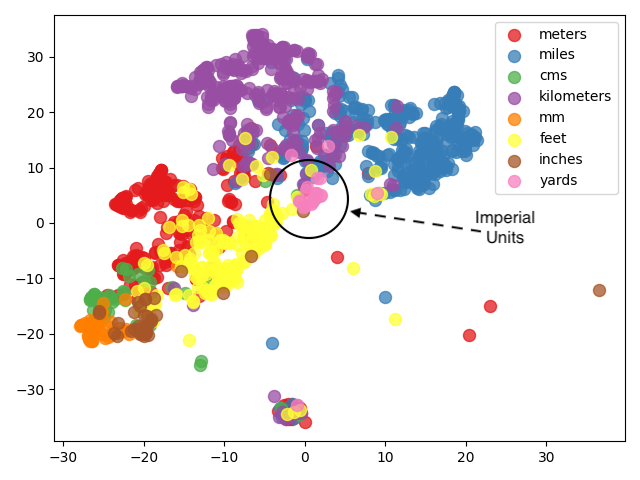}
    \hspace*{\fill}%
    \vspace{-2em}
    \caption{}
    \label{fig:u_tsne}
\end{subfigure}
\begin{subfigure}{.33\textwidth}
    \centering
    \includegraphics[width=1\textwidth]{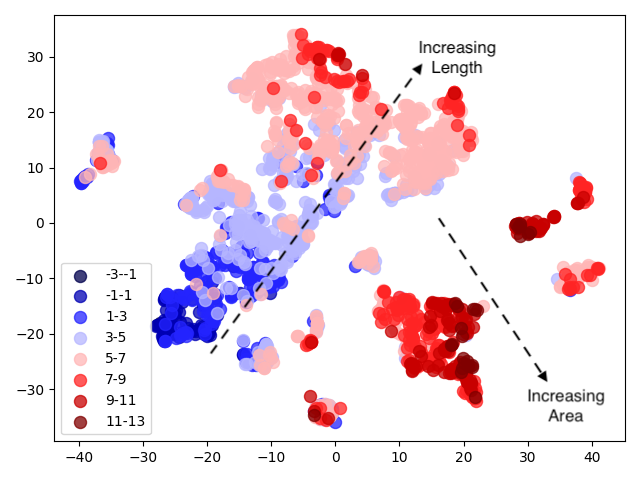}
    \hspace*{\fill}%
    \vspace{-2em}
    \caption{}
    \label{fig:x_tsne}
\end{subfigure}
\caption{t-SNE visualizations of semantic head embeddings labeled by \textbf{(left \ref{fig:d_tsne})} dimension, \textbf{(middle \ref{fig:u_tsne})} units of \emph{length}, and \textbf{(right \ref{fig:x_tsne})} number exponent bin. 
\textbf{Middle}: we observe a clustering of imperial units: feet, yards, miles.
\textbf{Right}: we show two directions where magnitudes of length and area measurements increase in value.}
\label{fig:tsne}
\end{figure*}
In Figure~\ref{fig:tsne} we plot the t-SNE embeddings of the sentences' $\hSem$, the output of our text encoder. We label each $\hSem$ with the masked measurement's true dimension, unit and exponent of the number.
In \ref{fig:d_tsne} we observe that most embeddings labeled by their true dimension tend to form tight clusters.
In \ref{fig:u_tsne} we filter to only show embeddings that share the \emph{length} dimension and label them by their units.
We find that clusters are organized by the relative magnitudes of their units
Kilometers and miles form the large cluster, feet and meters form the medium cluster, and millimeters, inches, and centimeters form the small cluster.
Further we see that \emph{yards} appear close to other \emph{imperial units} of \emph{feet} and \emph{miles}.
Finally, in \ref{fig:x_tsne} when embeddings are binned by the exponent of their values we observe that the left to right direction appears to capture the increasing magnitude of a number.

\section{Related Work}
\subsection{Numeracy}
Multiple works have probed word embeddings like word2vec, GloVe, FastText \cite{naik-etal-2019-exploring} and contextual embeddings from models like BERT \cite{Wallace2019Numbers, Zhang2020DoLE} or T5 \cite{Pal2021InvestigatingNL} on a variety of numerical tasks like sorting, numeration, magnitude prediction, and common sense \cite{Lin2020BirdsHF}.
Several works have targeted numeracy pretraining using left to right language models \cite{spithourakis2018numeracy}, CNN and RNN based models \cite{chen-etal-2019-numeracy}, pretrained transformers \cite{Spokoyny2020AnEI, Jin2021NumGPTIN}, for an overview \cite{thawani-etal-2021-representing}.

Incorporating synthetic mathematical data augmentations \cite{ggb2020injecting} has improved question answering \cite{Dua2019DROP} while numerical pretraining has been shown to lower masked language modelling perplexity \cite{thawani-etal-2021-numeracy}.
Either directly or indirectly units have been involved in providing more interpretable explanation of quantities \cite{chaganty-liang-2016-much}, solving Fermi problems \cite{Kalyan2021HowMC} and resolving numeric Fused-Heads \cite{Elazar2019WheresMH}.
\subsubsection{Numeracy Benchmarks}
Several numeracy benchmarks have been proposed like quantitative reasoning in natural language entailment \cite{ravichander2019equate} and synthetic measurement estimation \cite{Jin2021NumGPTIN}.

The closest benchmark to our work is the Distribution over Quantities dataset (DoQ), a large scale dataset of quantities introduced by
\citet{Elazar2019HowLA}. 
A rule-based method was combined with simple heuristics to build DoQ resulting in its high-coverage albeit also higher noise.
Although, Wiki-Convert is smaller, it has much higher fidelity since it utilizes a feature used by editors of Wikipedia to automatically convert quantities into different units.
Further, Wiki-Convert provides the whole sentence as context as opposed to triplets of words.
\citet{Zhang2020DoLE} use artificial templates to probe models on the DoQ dataset. They find little difference between numerically pretrained models and frozen embeddings such as ELMo. In contrast, our findings show there is a significant gap on Wiki-Convert between fully finetuned models and their frozen counterparts.

\section{Conclusion}
In this work we propose \cmpFull, a new task to resolve the limitation of  masked number prediction ($\mnp$) in which units are not considered.
In our study, we show probing of traditional pretrained transformers exposes a gap in their understanding of contextualized quantities.
Through careful quantitative and qualitative analysis of our new model, which directly reasons about underlying units and dimensions,
we find that it is possible to learn good representations of measurements. 
For future work we hope to extend this dataset to cover the thousands of existing standardized units from organizations such as UNECE.\footnote{United Nations Economic Commission for Europe}
We hope our $\cmp$ task encourages research into further development of better numeracy methodologies.

\bibliography{anthology,acl2020}
\bibliographystyle{acl_natbib}
\appendix
\section{Dataset}\label{app:data}

We train and evaluate our models on Wiki-Convert \cite{thawani-etal-2021-numeracy}, a dataset of English Wikipedia sentences where the number and unit in each sentence are human-annotated.
The built-in template in Wikipedia can ensure the text contains numbers and units.
For example, \texttt{\{\{convert|2|km|mi\}\}} displays as \texttt{2 kilometres (1.2 mi)}.
By searching within Wikipedia articles for the use of this template, the authors of Wiki-Convert automatically extract human-annotated numbers.
To perform unit canonicalization, we use Pint \footnote{Pint: https://github.com/hgrecco/pint} whenever the mapping is unambiguous.
In the ambiguous case, we manually inspect the sentence and perform the mapping.
For example, we map the unit \texttt{sqmi} in Wiki-Convert to \texttt{square miles} to let pint perform unit canonicalization.
Table \ref{tab:human-examples} shows examples of the extended dataset.
The original dataset contains 924,473 sentence.
The median sentence length is 106 characters, with 29,597 sentences has a length shorter than 20 characters. 
For preprocessing we exclude sentences which have more than 64 tokens to have efficient computing memory or where the number is negative for simplicity.

\section{MLM Preliminary Unit Probe}\label{app:mlm}
\begin{table*}
    \centering
    \resizebox{\textwidth}{!}{%
    \begin{tabular}{lccccccccccccc}
    \toprule

    \textbf{Input: [UNIT]} &m & km& ft& mi& yd& in& meters& kilometers& feet& miles& yards& inches& -
    \\
    \midrule
    \textbf{Output} &200& 10& 200& 2& 100& 1& 200& 20& 20& 2& 50& 3&-\\
    \midrule
    \textbf{Convertion factor}&1& 1000& 0.3048& 1609.34& 0.9144& 0.0254&1& 1000& 0.3048& 1609.34& 0.9144& 0.0254&-\\
    \midrule
    \textbf{Metric Output} &200.0&10000.0& 60.96& 3218.68& 91.44& 0.0254& 200.0& 20000.0& 6.096& 3218.68& 45.72& 0.0762&-\\
     \midrule
    \textbf{Mean (Metric Output)} & \multicolumn{12}{c}{-} &3086.8 m\\
    \midrule
    \textbf{std (Metric Output)} & \multicolumn{12}{c}{-} & 5820 m\\
    \bottomrule
    \end{tabular}%
    }
    \caption{Example outputs for \textbf{Alex Honnold climbed for [MASK] [UNIT]}.}
    \label{tab:output}
\end{table*}
We perform a preliminary unit probe with different unit inputs shown in Table  \ref{tab:output}. 
The model predicts vastly different numbers when conditioned on different units. 
We observe a mean of 3086.8 and a standard deviation of 5820 for all the converted metric output.

\section{Experiments}
\subsection{Quantitative Analysis}

\begin{figure}
\centering
  \includegraphics[width=.5\textwidth]{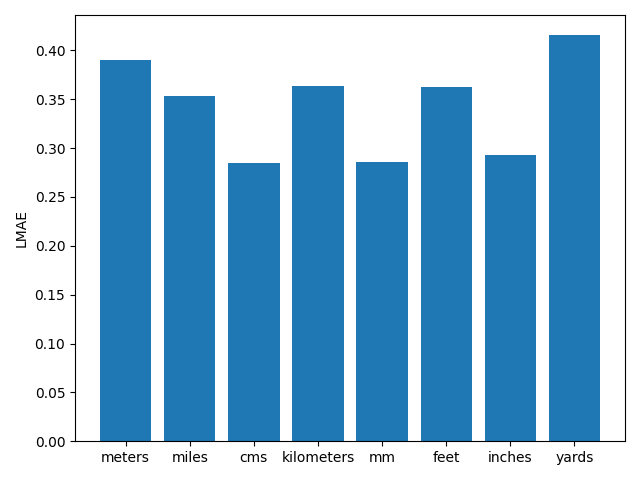}
    \caption{$\lmae$ $\downarrow$ by units of length. Predicting numbers for small magnitude units is easier than predicting numbers for their larger counterparts.}
  \label{fig:u_lmae}
\end{figure}

In Figure \ref{fig:u_lmae}, we show $\lmae$ is relatively small for small magnitude units, which means predicting numbers for small magnitude units is easier than predicting numbers for their larger counterparts.

\end{document}